\documentclass[lettersize,journal]{IEEEtran}
\usepackage{amsmath,amsfonts}
\usepackage{algorithmic}
\usepackage{algorithm}
\usepackage{array}
\usepackage[caption=false,font=normalsize,labelfont=sf,textfont=sf]{subfig}
\usepackage{textcomp}
\usepackage{stfloats}
\usepackage{url}
\usepackage{verbatim}
\usepackage{graphicx}
\usepackage{cite}
\usepackage{booktabs}
\usepackage[pagebackref,breaklinks,colorlinks]{hyperref}
\begin{document}

\title{Prototype-Aware Heterogeneous Task for Point Cloud Completion}

\author{Junshu~Tang, Jiachen~Xu, Jingyu~Gong, Haichuan~Song, Yuan~Xie, Lizhuang~Ma
\thanks{J. Tang, J. Xu, J. Gong and L. Ma are with the Department of Computer Science and Engineering, Shanghai Jiao Tong University, Shanghai 200240, China. E-mail: tangjs@sjtu.edu.cn, xujiachen@sjtu.edu.cn, gongjingyu@sjtu.edu.cn, ma-lz@cs.sjtu.edu.cn.}
\thanks{Haichuan~Song and Yuan~Xie are with the School of Computer Science and Technology, East China Normal University, Shanghai 200062, China. E-mail: hcsong@cs.ecnu.edu.cn, yxie@cs.ecnu.edu.cn.}}

\markboth{Journal of \LaTeX\ Class Files,~Vol.~14, No.~8, August~2021}%
{Shell \MakeLowercase{\textit{et al.}}: A Sample Article Using IEEEtran.cls for IEEE Journals}



\maketitle

\begin{abstract}
Point cloud completion, which aims at recovering original shape information from partial point clouds, has attracted attention on 3D vision community. Existing methods usually succeed in completion for \textit{standard} shape, while failing to generate local details of point clouds for some \textit{non-standard} shapes. To achieve desirable local details, guidance from global shape information is of critical importance. In this work, we design an effective way to distinguish \textit{standard/non-standard} shapes with the help of intra-class shape prototypical representation, which can be calculated by the proposed supervised shape clustering pretext task, resulting in a heterogeneous component w.r.t completion network. The representative prototype, defined as feature centroid of shape categories, can provide global shape guidance, which is referred to as soft-perceptual prior, to inject into downstream completion network by the desired selective perceptual feature fusion module in a multi-scale manner. Moreover, for effective training, we consider difficulty-based sampling strategy to encourage the network to pay more attention to some partial point clouds with fewer geometric information. Experimental results show that our method outperforms other state-of-the-art methods and has strong ability on completing complex geometric shapes.
\end{abstract}


\section{Introduction}
\label{sec:intro}
\IEEEPARstart{P}oint clouds, obtained by LiDAR sensors or randomly sampled from 3D CAD models, have increasingly been used to represent 3D scenes or objects. However, due to the occlusion and insufficient lighting, obtained point clouds always have distinct missing regions, which will impact the representation of the object. To deal with this problem, point cloud completion is required to reconstruct complete geometric shapes using parts of point clouds.

\begin{figure}[ht]
    \centering
    \includegraphics[width=\linewidth]{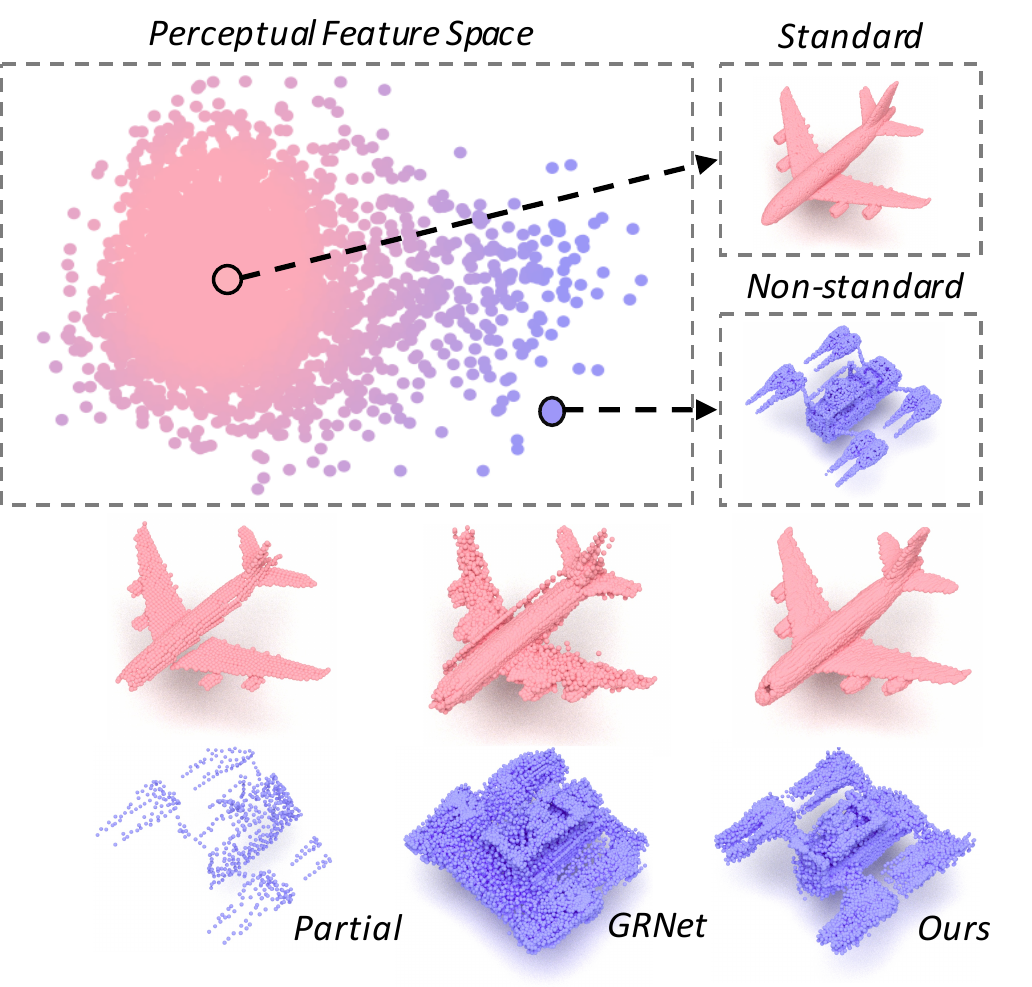}
    \caption{Illustration of the perceptual feature space and point cloud completion on standard (Red) and non-standard (Blue) shape of airplane.
    Compared with GRNet\protect\cite{xie2020grnet}, our method focus more on predicting local and complex geometric details of non-standard samples.} 
    \label{fig:fig1}
\end{figure}

Previous methods made impressive progress in point cloud completion ~\cite{yuan2018pcn,tchapmi2019topnet,huang2020pf,sarmad2019rl,wang2020cascaded, xiang2021snowflakenet, yu2021pointr,tang2022lake}. However, these methods usually succeed to complete the object with the regular geometric shape ({\it i.e.,} standard shape), and fail to predict local details and complex geometry of the object which is quite unusual ({\it i.e.,} non-standard shape). Taking airplane as an example, as shown in the lower part of Fig.~\ref{fig:fig1}, previous methods, such as GRNet~\cite{xie2020grnet}, do well in completing airplanes with long fuselage, wings, engines, tail fin and horizontal stabilizer (marked in red). When it comes to some specific shapes (marked in blue), these methods tend to predict mean shape with blurry details.

To further explore the reason, we employ a feature extractor which embeds complete point clouds into a latent feature space with geometric information, also named as perceptual feature space. We exploit the feature distribution for each category. Specifically, we build the feature extractor based on PointNet++~\cite{qi2017pointnet++}, and train it to achieve supervised object classification. In the upper part of Fig.~\ref{fig:fig1}, we visualize the feature distribution for the airplane category in the 2D plane. We can observe that the standard shape is similar and accounts for a large portion (geometric features converge in a small scope) while the non-standard shape is diverse and accounts for a small portion (their geometric features diverge around). Based on this observation, the main reason why previous works failed to complete non-standard shapes is that, the network trained on imbalanced dataset tends to overfit standard shapes with large samples. Similarly, previous works focused on the long-tailed problem caused by imbalanced dataset in many fields such as face recognition~\cite{cao2020domain, meng2021magface} and 
image/video classification~\cite{zhu2020inflated}. Compared with these tasks with exclusive labels, in point cloud completion, the shape imbalance exists within each category, so it is hard to distinguish ``head" shapes with a large portion (standard) and ``tail" shapes with a small portion (non-standard). 

To tackle this problem, we propose a prototype-aware heterogeneous task for point cloud completion.
First, to distinguish the standard and non-standard shape for each category, we utilize the Gaussian Mixture Model (GMM) to fit the latent space embedded by the feature extractor and find the prototypical feature center. Then, we define the soft-perceptual prior as the distance between the center and each latent feature to provide useful global shape guidance for point cloud completion. We take this component as a supervised shape clustering pretext task or heterogeneous task, which has different objectives but involves complicated interactions with subsequent point cloud completion task.

To effectively integrate the soft-perceptual prior in completion network, we propose a \textbf{S}elective \textbf{P}erceptual feature \textbf{F}usion (SPF) module. It can fuse the prior with global and local features in a multi-scale manner for the basic encoder-decoder architecture. Compared with the skip-attention module in ~\cite{wen2020point}, our SPF module guided by global geometric information could utilize the global and local features in a more effective way.

{ \color{black}
Additionally, due to the diversity of environments,
the missing part of point clouds varies a lot. Some partial point clouds miss negligible parts, while others miss indispensable parts with important geometric information, resulting in ``easy" and ``hard" samples, respectively. Therefore, we propose a difficult-sampling mechanism following the principle that training samples should not be treated equivalently. Some partial point clouds will need more attention if their global shapes are more difficult to identify. To this end, we calculate the distance between partial input and the corresponding complete point cloud in the perceptual feature space. This distance will be adopted to rank the importance of samples and reweight the loss term for training.}

In general, the main contributions of our method are summarized as follows:
\begin{itemize}
    \item A supervised shape clustering pretext task is proposed to obtain shape prototypes and further generate shape guidance, namely soft-perceptual prior, for downstream point cloud completion network. 
    \item We propose a selective perceptual feature fusion module to fuse the soft-perceptual prior of partial point cloud with both global and local features in a basic point cloud completion network.
    \item We construct a novel strategy to rank the importance of each sample based on the perceptual feature gap and apply a difficulty-based sampling scheme which strengthens the attention on samples with fewer geometric information.
    \item Experimental results show that our method outperforms existing methods on most of the sample point clouds especially some non-standard samples.
\end{itemize}

\begin{figure*}[htb]
    \centering
    \includegraphics[width=\linewidth]{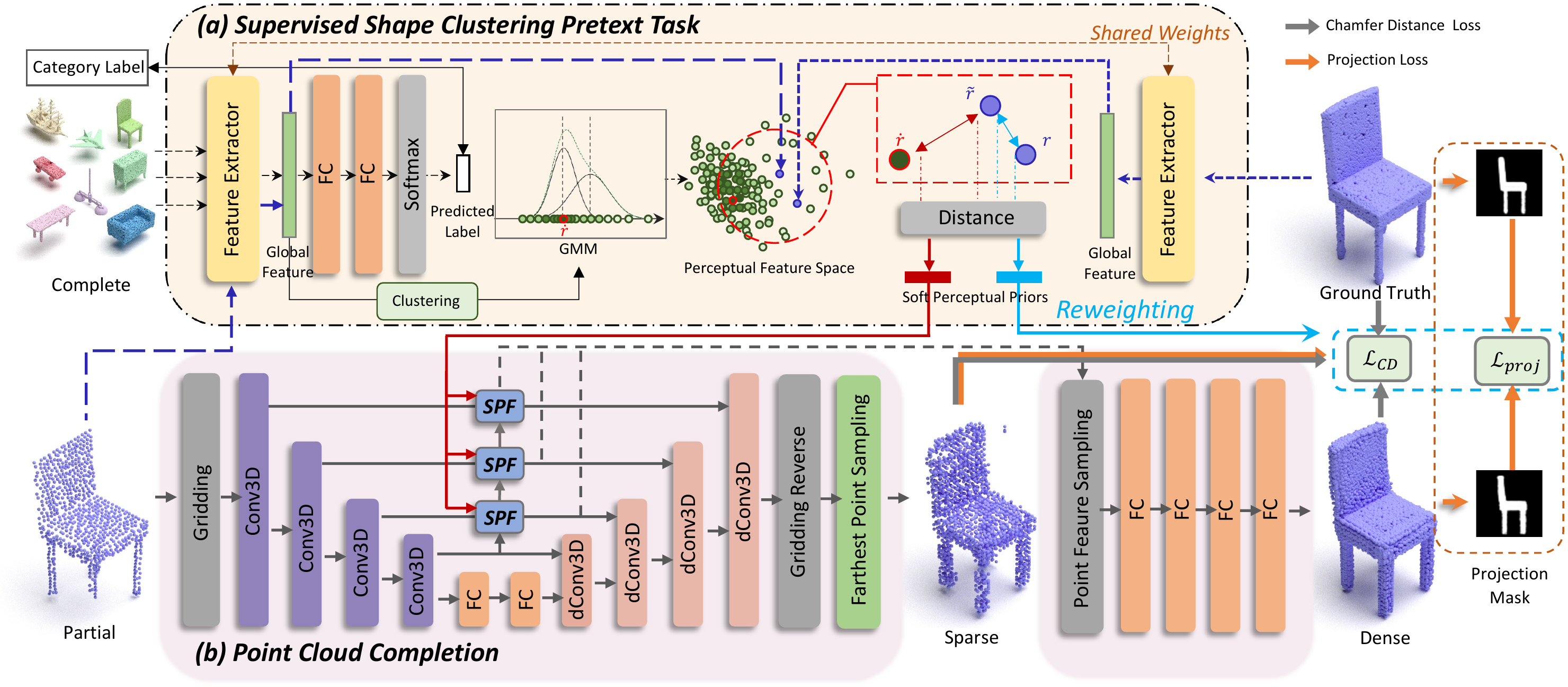}
    \caption{The overall architecture of our network. (a) represents the supervised shape clustering pretext task which learns soft-perceptual priors of each point cloud. (b) represents the point cloud completion baseline. \textit{SPF} refers to our proposed selective perceptual feature fusion module. \textit{FC} refers to the fully connected layer. We apply the \textit{Gridding}, \textit{Gridding Reverse} and \textit{Point Feature Sampling} operation from GRNet.} 
    \label{fig:overall-architecture}
\end{figure*}

\section{Related Work}

\subsection{Point Cloud Completion}
Point cloud completion approaches aim to recover geometric shapes from partial inputs which have attracted significant interest \cite{yuan2018pcn, tchapmi2019topnet, zhang2020detail, huang2020pf, xie2020grnet, pan2021variational, Wen_2021_CVPR}.  
Recently, inspired by deep learning based point cloud processing approaches \cite{qi2017pointnet, qi2017pointnet++}, researches of deep learning based point cloud completion made great progress. PCN~\cite{yuan2018pcn} firstly utilized the shared Multi-Layer Perceptrons (MLPs) and encoder-decoder architecture to process the partial input, generate intermediate coarse point clouds, and then output dense point cloud with more local details. 
TopNet~\cite{tchapmi2019topnet} proposed a tree-based decoder to model the topology structure of geometric shapes. 
Following these work, NSFA~\cite{zhang2020detail} extracted multi-level features and reconstructed known and missing parts with local and global representations respectively, and then also adopted a coarse-to-fine decoding process which constrains the densification of point clouds progressively. 
PFNet~\cite{huang2020pf} further utilized a multi-resolution encoder and a point pyramid decoder for point cloud generation, and apply discriminators for optimization. VRCNet
~\cite{pan2021variational} also designed multi-scale local point feature fusion module to enhance structural relations for point cloud analysis. 
In addition to these point-based methods, GRNet~\cite{xie2020grnet} tended to use 3D fixed grids as the intermediate representation of point clouds, and use 3D convolution network (3DCNN) to process them, which explicitly preserves the structural and context of point clouds and improve the performance of completion. We also applied this voxelization process as our backbone for point cloud completion.

\subsection{Self-supervised Learning for Point Clouds}
Recently, several approaches utilized self-supervised tasks for learning better representations for point cloud analysis mostly in a pre-training stage. Similar to ~\cite{noroozi2016unsupervised} which solved Jigsaw Puzzles in 2D images, ~\cite{sauder2019self} reassemble raw point clouds whose parts have been randomly replaced. The pretext task is learned to find the voxel assignment that reconstructs the original point cloud. ~\cite{thabet2019mortonnet} learned a predict-the-next-point self-supervised task to generate Z-ordered sequences from an unstructured point cloud. 
~\cite{zhang2019unsupervised} and ~\cite{hassani2019unsupervised}
trained networks for multi-pretext-tasks, such as clustering or reconstruction from raw inputs. In addition, ~\cite{chen2019deep} predicted pairwise relationships by auto-encoder in various input points for downstream point cloud analysis. ~\cite{tang2020improving} learned local geometric features by training a network to predict the normal vector and curvature. 
Compared to these studies, our work learns a prototypical-aware heterogeneous task that extracts compact representations of point clouds in different categories. And we utilize deep clustering method for prototypical representation learning which provides a global guidance for point cloud learning. It is noteworthy that, our proposed supervised shape clustering task shares a similar methodology with self-supervised learning, where the useful knowledge is transferred from relative simple tasks on auxiliary datasets. Our proposed method aims to construct global shape prior that is semantically meaningful via supervised shape clustering pretext task, to facilitate the downstream completion task. However, different from common self-supervision methods, we utilize the available complete point cloud in the training set as the input of pretext task to learn the prototype for each category more precisely. 

\subsection{Prototypical Representation Learning}

Prototypical representation learning is recently popular in few-shot learning~\cite{wang2020prototypical,zhu2020robust,yue2021prototypical, pahde2021multimodal,NightCity} and unsupervised learning~\cite{li2020prototypical,zhang2021prototypical}. A prototype is defined as the centroid of feature embedding space which represents the semantic structure. Samples that are matched with the same prototypical center have similar semantic information. Therefore, prototypical  representation learning provides a kind of soft pseudo-label that represents the relationship among instances according to different feature distances to prototypical centers. The prototype could be learned online or offline. ~\cite{zhang2021prototypical} proposed to update each prototypical center on-the-fly throughout the training, while ~\cite{li2020prototypical} and ~\cite{yue2021prototypical} applied deep clustering method to learn centers in different domains in an offline way. The prototypical center is learned by Expectation-Maximization (EM) based algorithm. Our proposed method also utilize clustering method and EM algorithm to learn the prototypical center in each shape category. Specifically, our method utilizes the prototypical center to capture fine-grained intra-class semantic similarity without supervision.



\section{Method}
In this paper, we propose an architecture for point cloud completion by constructing a supervised shape clustering pretext task. The overall architecture of our method is illustrated in Fig.\ref{fig:overall-architecture}. It consists of a two-stage process: the supervised shape clustering pretext task and point cloud completion. 
First, in the Fig.\ref{fig:overall-architecture}(a), given a set of complete point clouds with corresponding category labels, we utilize a pre-trained feature extractor to encode the raw point cloud into a latent feature space. To exploit the feature distribution of each category, we fit GMM distributions to the subspace spanned by all features belonging to the same category. A modified Expectation Maximization (EM) algorithm is derived to estimate the parameters and find prototypical center $\dot{r}$ in each subspace. We save the prototypical center of each category to guide the downstream completion pipeline.

Second, the pipeline in point cloud completion (Fig.\ref{fig:overall-architecture}(b)) is that, given a partial input $\tilde{P}$, we firstly transform the partial point cloud into 3D grids, and then use 3DCNN based architecture to generate sparse point cloud $\hat{P}^{sp}$. After that, the sparse point cloud and features from previous layers are input to MLPs to predict the final dense point cloud $\hat{P}$. 
To make the previous heterogeneous task guide the completion process, 
we use the pre-trained feature extractor to extract the partial feature $\tilde{r}$, whose nearest prototypical center $\dot{r}_j$ from subspace of $j$-th category will be found.

\begin{figure}[]
    \centering
    \includegraphics[width=\linewidth]{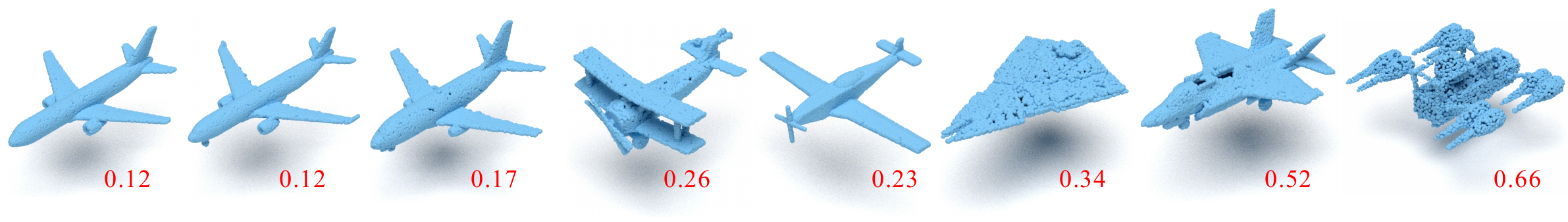}
    \label{fig:2}
    \caption{Illustration of sampled geometries in Airplane category pairs with different missing part and the euclidean distance to the prototypical center of corresponding class. The results illustrate that samples with similar shapes have similar distance while non-standard shapes are further from the center.}
\end{figure}

The distance between $\tilde{r}$ and $\dot{r}_j$ is regarded as the soft-perceptual prior of partial input. We set the pretext task as a classification task with only coordinate input to encode geometry information into the latent space. Therefore, the euclidean distance can represent the geometric difference. We also list some geometries and the euclidean distance to the prototypical center which is shown on Fig.~\ref{fig:2}. The experimental results show that samples with similar shapes have similar distance while non-standard shapes are further from the center.

To incorporate the prior in our framework, we design an 
SPF module to plug into encoder-decoder architecture for manipulating the feature aggregation. Additionally, to emphasize some non-standard samples, particularly for some  point clouds with less shape information, the reweighted loss is constructed with the help of difficult-sampling mechanism. In this way, the hard sample mining can be realized.

\begin{figure}
    \centering
    \includegraphics[width=\linewidth]{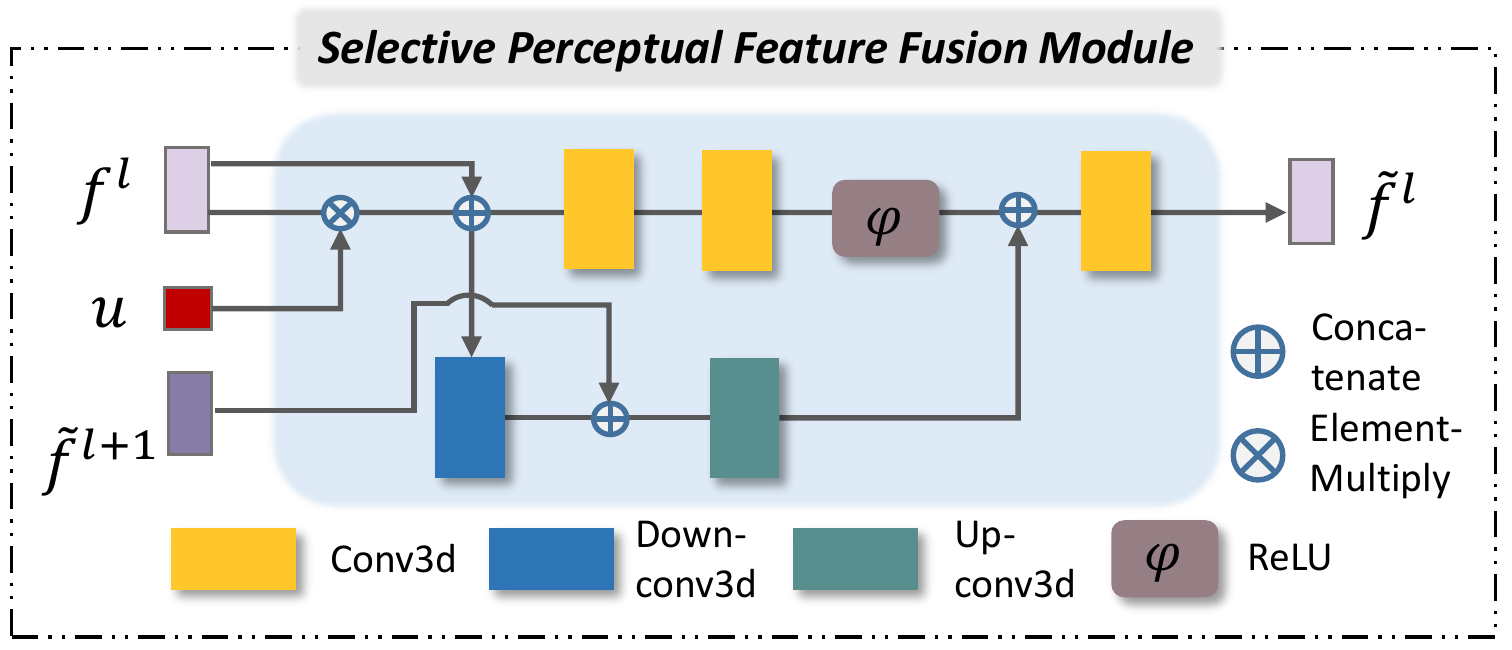}
    \caption{The detail structure of SPF module at layer $l$.} 
    \label{fig:spf}
\end{figure}

\subsection{Soft-perceptual Priors Learning}
\label{subsec:priornet}
Prior is widely used in 3D tasks, especially in reconstruction problems~\cite{pavlakos2019expressive,sun2020disp,wang2020cascaded}. 
To learn the prior in a data-driven manner, the common choice is the Variational AutoEncoder (VAE) or its derivation~\cite{pavlakos2019expressive, sun2020disp}, which encodes all input data into a latent space which represents the shape information. Most shape priors are regularized by a Gaussian distribution. The main reason is that the average of a large number of independent and identically distributed(i.i.d) random variables is approximately Gaussian distributed according to the central limited theory. In our paper, we also learned priors based on the assumption that latent  intra-class representations are Gaussian distributed.  However, in the completion task, various categories of partial objects need to be completed. 
Since the huge gap exists across different objects, constructing an independent VAE for each category is rather costly.

Instead, we introduce the supervised shape clustering as the pretext task to provide soft-perceptual prior for downstream completion, deducing roughly global shape information for partial point clouds. After pre-trained feature extractor, we take the output global feature as the latent representation $r$ in feature space, where features of objects in similar shape are expected to have similar representations. Therefore, all features of the same category span a subspace. It is noteworthy that in this subspace, for some non-standard objects, although they are classified into correct categories, their features have significant difference compared with other objects belonging to the same category. To this end, we aim to find a prototype, which could represent a standard shape, by calculating a density center in each category subspace. Specifically, we utilize a GMM to fit the feature distribution of each subspace: where the EM algorithm is used to estimate $\pi_{k}, \mu_k, $ and $\sigma_{k}$, and find cluster centers.
To balance the computational complexity and the performance, we set $K=4$, and run the algorithm for 20 iterations.  Considering that features of standard shapes always bunch up while other features diverge around, we calculate the prototypical center $\dot{r}$ according to the corresponding density in each cluster. {\color{black} It is noteworthy that the goal of our proposed clustering task is to capture fine-grained intra-class semantic information and distinguish whether the sample is standard or not. Samples in smaller portion are assigned more attention which will not go against our motivation.} For subsequent calculations, we save both $\dot{r}$ and the maximum radius $\gamma$ for each category. The detail of this algorithm is shown as follows.

Specifically, we save prototypical feature center $\dot{r}$ and the maximum radius $\gamma$ for each categories for the following completion task. Take an explicit class $\mathcal{C}$ as an example, we define $N_\mathcal{C}$ as the number of point clouds. The pseudo code of this algorithm is presented in Algorithm \ref{alg:A}:

\begin{algorithm}[t]
\caption{Find prototypical feature center $\dot{r}$ for class $\mathcal{C}$}
\label{alg:A}
\begin{algorithmic}
\STATE \textbf{Input:} A PriorNet $\Psi$ to extract global feature $r_{i}$ of a complete point cloud $P_{i}$, $i\in[1,N_\mathcal{C}]$.
\STATE \textbf{Input:} The number of distributions $K$.
\STATE \textbf{Initialize:} The parameters $\pi_k,\mu_{k},\sigma_{k}, k\in[1,K]$.
\FOR{iteration = $1, E$}
\STATE $\textbf{E}$-step: Each $r_{i}$ gets responsibility $d_{k,i}$, which varies between clusters. $d_{k,i}=\frac{\pi_k\mathcal{N}(r_{i};\mu_k,\sigma_k)}{\sum_{k'}\pi_{k'}\mathcal{N}(r_{i};\mu_{k'}, \sigma_{k'})}$. 
\STATE $\textbf{M}$-step: Update cluster centers: $r_{k}=\frac{\sum_{i}r_{i}d_{k,i}}{\sum_{i}d_{k,i}}$, 
\STATE Update parameters $\pi_k$, $\mu_k$ and $\sigma_{k}$: 
\STATE$\mu_{k}=\frac{1}{N_k}\sum_{i}d_{k,i}r_{i}$,\\ $\sigma_k=\frac{1}{N_k}\sum_{i}d_{k,i}(r_{i}-\mu_k)(r_{i}-\mu_k)^{T}$,
\STATE where $N_k=\sum_{i}d_{k,i}$ and $\pi_k=N_k/N_\mathcal{C}$.
\ENDFOR
\STATE Set $K=4$ and $E=20$.
\STATE Final assignment $S_{i} = \{ x_{p}:d_{i,p}\leq d_{j,p} \forall j, 1\leq j\leq K\}$
\STATE Set dense center $\dot{r}=(\sum_{i}|S_{i}|r_{i}) / (\sum_{i}|S_{i}|)$,
\STATE\textbf{Output:} The dense center $\dot{r}_\mathcal{C}$ as prototypical feature center of the category $\mathcal{C}$.
\STATE \textbf{Output:} The maximum euclidean radius $\gamma_\mathcal{C}= \max\{||r_{i}-\dot{r}_\mathcal{C}||_2,i\in[1,N_\mathcal{C}]\}$ of the category $\mathcal{C}$.
\end{algorithmic}
\end{algorithm}

\begin{figure*}[t]
    \centering
    \includegraphics[width=\linewidth]{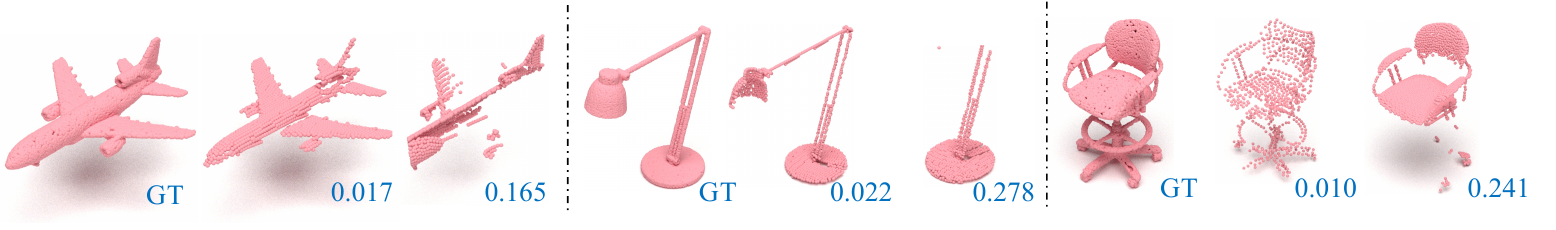}
    \caption{Illustration of sampled partial-complete pairs with different missing part and feature cosine distance. We can see that different partial objects results in different levels of information loss. A larger cosine distance indicates more missing semantic information.}
    \label{fig:distance}
\end{figure*}

\subsection{Selective Perceptual Feature Fusion Module}

Given a set of $N$ partial point clouds $\{\tilde{P}_i|i=1,...,N\}$, we utilize the pre-trained feature extractor $\Psi$ 
to extract the latent representation $\tilde{r}_i = \Psi(\tilde{P}_i)$ for each partial point cloud. Then we calculate the Euclidean distance $u_i$ between the representation $\tilde{r}_i$ and its nearest prototypical center $\dot{r}_j$:
\begin{equation}
\label{equation:prototypical}
\begin{aligned}
    &u_i=\min\{u^{1}_i, u^2_i,\cdots, u_i^C\}, \\
    &u^{j}_i = (||\tilde{r}_i-\dot{r}_j||_2)/\gamma_j, j\in[1, C],
\end{aligned}
\end{equation} 
where $j$ denotes the category label and $C$ represents the number of categories. {\color{black} We assume a standard normal distribution to improve the efficiency where 2D Gaussian will not be tall and over-shaped, so we use $\gamma_j$ as the maximum radius for $j$-th category.} Consequently, $u_i$ is regarded as the soft-perceptual prior of partial point cloud $\tilde{P}_i$, implicitly containing information of prototypical shape.

To effectively integrate the soft-perceptual prior, we construct a Selective Perceptual feature Fusion module (SPF) to fuse priors in a manner of layer-by-layer conduction. This mechanism encourages SPF to preserve the consistency of the structure information in different resolutions. We apply this module on a GRNet embedded 3DCNN U-Net pipeline, where the direct skip connection is replaced by our SPF, as shown in Fig. \ref{fig:overall-architecture}(b).

The detail of the SPF module is described in Fig.~\ref{fig:spf}.Taking SPF module in the $l$-th layer as an example, we obtain the input feature $f^{l}$ from the $l$-th layer of encoder and $\tilde{f}^{l+1}$ from SPF in the
$(l+1)$-th layer. 
The $f^l$ is multiplied with soft-perceptual prior $u$, and is down-sampled by convolution block to be concatenated with high-level feature $\tilde{f}^{l+1}$. After that, an up-sampling convolution block transforms the merged feature to $l$-th level. At last, we concatenate two feature maps and use another convolution unit to output the fusion feature $\tilde{f}^{l}$ which will be is also 
used for the SPF in $(l-1)$-th layer. Therefore, the network is able to aggregate features collaboratively in multi-scale and selectively for each object.

\subsection{Difficulty-based Sampling}

{\color{black}In traditional point cloud completion approaches, all partial-complete pairs are considered as equally important and share the same weights for training.
However, there are various levels of information loss on different partial objects (as shown in Fig~\ref{fig:distance}). To perform hard sample mining, we propose a difficulty-based sampling strategy with the help of prototypical representations. Specifically, we weight the partial sample differently during training. With this strategy, the training process will pay more attention to hard samples other than treating all ones equivalently.}
Given a pair of complete and partial point clouds $P_i$ and $\tilde{P}_i$, we extract their latent representations $r_i=\Psi(P_i)$ and $\tilde{r}_i=\Psi(\tilde{P}_i)$. Then, we evaluate the distance between $r_i$ and $\tilde{r}_i$ by Cosine Similarity Metric:
\begin{equation}
    \begin{aligned}
    cos(r_i, \tilde{r}_i) &= \frac{r_i\cdot\tilde{r}_i}{||r_i||\cdot||\tilde{r}_i||}.\\ 
    \end{aligned}
\end{equation}

To focus more on hard samples, which suffer from shape information loss more seriously and are harder to be completed, we utilize the calculated metric and reweight the loss term (see Eq.(\ref{loss-func}) below). 
Specifically, we define the weight $\textbf{w}_{i}$ for partial point cloud $\tilde{P}_i$ as follows:
\begin{equation}
    \begin{aligned}
        \textbf{w}_{i} = 1+\sigma(1-cos(r_i, \tilde{r}_i)),
    \end{aligned}
\end{equation}

where $\sigma(x) = 1/(1 + \exp^{k(t-x)})$ is an activation function. We set the threshold $t=0.25$ and $k=8$.

\begin{figure*}
    \centering
    \includegraphics[width=\linewidth]{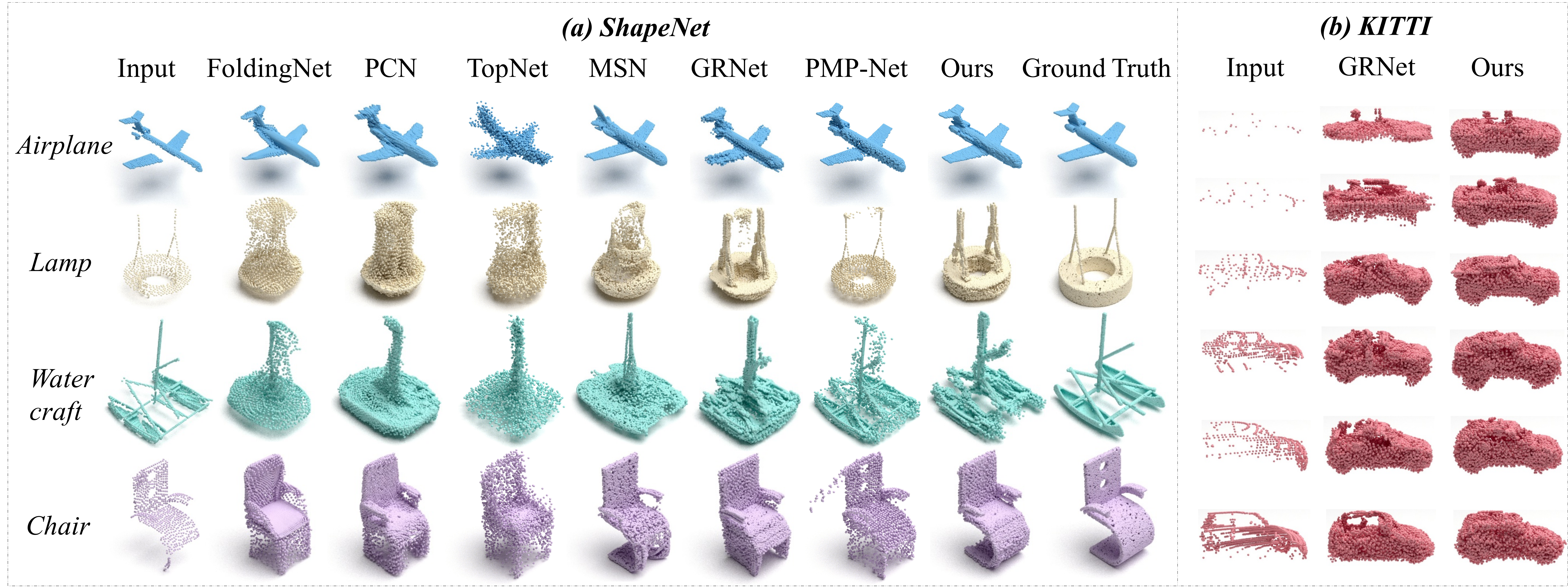}
    \caption{Illustration of point cloud completion comparison with previous methods on different categories of ShapeNet dataset (a) and KITTI dataset (b). For ShapeNet, from left to right columns are the partial input, completed results from FoldingNet, PCN, TopNet, MSN, GRNet, PMP-Net, our method and the last is the ground truth. Illustration for KITTI includes the partial input and completed results from GRNet and our method.}
    \label{fig:compare}
\end{figure*}

\subsection{Training Objective}
First, we apply the Chamfer Distance (CD) loss on both sparse and dense point clouds, following most of methods for completion~\cite{yuan2018pcn,tchapmi2019topnet,xie2020grnet}.
Given the predicted point cloud $\hat{P}$ and the ground truth $P$, the CD loss is defined as follows:
\begin{equation}
    \begin{aligned}
    \mathcal{L}_{CD}(\hat{P},P) =\ \frac{1}{|\hat{P}|}\sum_{a\in \hat{P}}\min_{b\in P}||a-b||^{2} + \frac{1}{|P|}\sum_{b\in P}\min_{a\in \hat{P}}||b-a||^{2}.
    \end{aligned}
\end{equation}

Besides the CD loss, we apply a projection constraint under multi-views to ensure the predicted point clouds have similar silhouette information as the ground truth. By adopting a differentiable rendering method proposed in ~\cite{navaneet2019capnet}, given a predicted point cloud $\hat{P}$, 
we randomly sample $M$ points and orthogonally project $\hat{P}$ with a viewpoint $v$ into a mask $\hat{\mathbf{M}}^{v}\in \mathbb{R}^{H\times W}$, where $H$ and $W$ represent the height and width of the mask image, respectively. Specifically, the pixel value on the mask image 
$\hat{\mathbf{M}}^{v}_{h,w}$ is calculated as:
\begin{equation}
    \begin{aligned}
    \hat{\mathbf{M}}^{v}_{h,w} = \tanh\left(\sum_{m=1}^{M}\phi(\hat{x}_m-h)\phi(\hat{y}_m-w)\right),
    \end{aligned}
\end{equation}
where $\phi(\cdot)$ is a Gaussian kernel. $\hat{x}_{m},\hat{y}_m$ and $h,w$ denote the orthogonal and pixel coordinates, respectively. We also render the completed ground truth $P$ with the same viewpoint as well into the mask image $\mathbf{M}^{v}$. Then, we define the projection loss $\mathcal{L}_{proj}$ as the difference between $\mathbf{M}^{v}$ and $\hat{\mathbf{M}}^{v}$ using the binary cross-entropy loss:
\begin{equation}
    \begin{aligned}
    \mathcal{L}_{proj} = -\frac{1}{HWV}\sum_{v}^{V} \sum_{h,w}^{H,W}&(\hat{\mathbf{M}}^{v}_{h,w}log(\mathbf{M}^{v}_{h,w}+\epsilon) \\&+ (1-\hat{\mathbf{M}}^{v}_{h,w})log(1-\mathbf{M}^{v}_{h,w}-\epsilon)),
    \end{aligned}
\end{equation}
where $V$ denotes the number of viewpoints $v$, and $\epsilon$ is a tiny tolerance to prevent invalid numerical operations. In our work, we set $M=512, H=W=64, V=8,$ and $\epsilon=1e-8$. By considering the difficulty-based sampling mechanism, the total loss for the network is as follows:
\begin{equation}
    \begin{aligned}
        \mathcal{L} = \sum_{b=1}^{B} \textbf{w}_{b}( &\mathcal{L}_{CD}(\hat{P}^{sp}, P) +\mathcal{L}_{CD}(\hat{P}, P)\\ & +\lambda_{proj}(\mathcal{L}_{proj}(\hat{P}^{sp}, P) + \mathcal{L}_{proj}(\hat{P}, P)))),
    \end{aligned}
    \label{loss-func}
\end{equation}
where $B$ denotes the batchsize of each iteration. $\hat{P}^{sp}$ and $\hat{P}$ represent the predicted sparse and dense point cloud, respectively.

\section{Experiments}
\subsection{Dataset Setting}
\textbf{ShapeNet:} The ShapeNet~\cite{chang2015shapenet} dataset for point cloud completion is created by PCN~\cite{yuan2018pcn}, including different objects from 8 categories: airplane, cabinet, car, chair, lamp, sofa, table, and vessel. The training set contains 28,974 objects, while validation and test set contains 800 and 1,200 objects, respectively.
The complete point cloud consists of 16,384 points which are uniformly sampled on the original CAD model. Partial point cloud, consisting of 2,048 points, is created by back-projecting 2.5D depth images into 3D from 8 random viewpoints.

\noindent\textbf{Completion3D:} 
Completion3D\footnote{\url{http://completion3d.stanford.edu/}} is a widely-used benchmark for point cloud completion, which is firstly created on TopNet~\cite{tchapmi2019topnet}. The testing set includes 1,184 partial point clouds without the ground truth. Different from ShapeNet dataset, the number of points for evaluation in Completion3D is 2,048.

\noindent\textbf{KITTI:} KITTI dataset for point cloud completion, which is also created by PCN, consists of 2,483 partial point clouds with corresponding bounding boxes and track labels from real-word LiDAR scans in 425 world frames. It is noteworthy that we randomly select 2,048 points for samples which have more than 2,048 points.

\subsection{Evaluaion Metrics}

We utilize two evaluation metrics between output dense point cloud and the ground truth, Chamfer Distance and F-Score, following most of the methods~\cite{tatarchenko2019single,xie2020grnet} on ShapeNet test set. Since we don't have the ground truth of Completion3D, we submit the completed results to its server for testing. 
{\color{black}We denote} $\hat{P}$ as the predicted point cloud and $P$ as the ground truth. In this section, we introduce the computational formula of each metrics. 

\noindent\textbf{Chamfer Distance.} The Chamfer Distance metric between $\hat{P}$ and $P$ is defined as:
\begin{equation}
    \begin{aligned}
    CD(\hat{P},P) = \frac{1}{|\hat{P}|}\sum_{a\in \hat{P}}\min_{b\in P}||a-b||^{2} + \frac{1}{|P|}\sum_{b\in P}\min_{a\in \hat{P}}||b-a||^{2}.
    \end{aligned}
\end{equation}

\noindent\textbf{F-Score.} F-Score metric can also evaluate the performance of point cloud completion. The F-Score between $\hat{P}$ and $P$ is defined as:
\begin{equation}
    \begin{aligned}
        F\text{-}Score(\sigma) = \frac{2G(\sigma)H(\sigma)}{G(\sigma)+H(\sigma)},
    \end{aligned}
\end{equation}
where $G(\sigma)$ and $H(\sigma)$ denote the point-wise precision and recall for a threshold $\sigma$, respectively. In our work, we set threshold $\sigma=0.01$.
\begin{equation}
    \begin{aligned}
        G(\sigma)=\frac{1}{|\hat{P}|}\sum_{a\in\hat{P}}\left[\min_{b\in P}||b-a||<\sigma\right], \\ H(\sigma)=\frac{1}{|P|}\sum_{b\in P}\left[\min_{a\in \hat{P}}||a-b||<\sigma\right].
    \end{aligned}
\end{equation}


\begin{table}[]
\setlength{\tabcolsep}{0.6mm}
\centering\scriptsize
\begin{tabular}{l|cccccccc|c}

\toprule
CD$(\times10^{-3})$$\downarrow$& Airplane & Cabinet & Car & Chair & Lamp & Sofa & Table & Vessel & Overall \\ \midrule
Folding\protect~\cite{yang2018foldingnet} & 9.49 & 15.80 & 12.61 & 15.55 & 16.41 & 15.97 & 13.65 & 14.99 & 14.31\\
PCN\protect~\cite{yuan2018pcn} & 5.50 & 22.70 & 10.63 & 8.70 & 11.00 & 11.34 & 11.68 & 8.59 & 9.64\\
AtlasNet\protect~\cite{groueix2018papier} & 6.37 & 11.94 & 10.10 & 12.06 & 12.37 & 12.99 & 10.33 & 10.61 & 10.85 \\
TopNet\protect~\cite{tchapmi2019topnet} & 7.61 & 13.31 & 10.90 & 13.82 & 14.44 & 14.78 & 11.22 & 11.12 & 12.15 \\
MSN\protect~\cite{liu2020morphing} & \textbf{4.85} & 11.00 & 8.99 & 9.16 & 10.34 & 10.68 & 8.35 & 9.24 & 9.07 \\
GRNet\protect~\cite{xie2020grnet} & 6.45 & 10.37 & 9.45 & 9.41 & 7.96 & 10.51 & 8.44 & 8.04 & 8.83\\
PMP-Net\protect~\cite{wen2021pmp} & 5.50 & 11.10 & 9.62 & 9.47 & \textbf{6.89} & 10.74 & 8.77 & \textbf{7.19} & 8.66\\\midrule
Ours & 6.09 & \textbf{9.22} & \textbf{8.86} & \textbf{8.55} & 7.25 & \textbf{9.33} & \textbf{7.64} & 7.53 & \textbf{8.05} \\ \bottomrule
\end{tabular}

\caption{Quantitative comparison results with other methods of point cloud completion on ShapeNet using per-point L1 Chamfer Distance (CD, lower is better). Note that the metric is computed by 16,384 points.}
\label{tab:shapenetcd}
\end{table}

\begin{table}[]
\setlength{\tabcolsep}{0.6mm}
\centering\scriptsize
\begin{tabular}{l|cccccccc|c}

\toprule
F-Score$\uparrow$ & Airplane & Cabinet & Car & Chair & Lamp & Sofa & Table & Vessel & Overall \\ \midrule
Folding\protect~\cite{yang2018foldingnet} & 0.642 & 0.237 & 0.382 & 0.236 & 0.219 & 0.197 & 0.361 & 0.299 & 0.322\\
PCN\protect~\cite{yuan2018pcn} & 0.881 & \textbf{0.651} & \textbf{0.725} & 0.625 & 0.638 & 0.581 & 0.765 & 0.697 & 0.695\\
AtlasNet\protect~\cite{groueix2018papier} &  0.845 & 0.552 & 0.630 & 0.552 & 0.565 & 0.500 & 0.660 & 0.624 & 0.616 \\
TopNet\protect~\cite{tchapmi2019topnet} & 0.771 & 0.404 & 0.544 & 0.413 & 0.408 & 0.350 & 0.572 & 0.560 & 0.503  \\
MSN\protect~\cite{liu2020morphing} & 0.854 & 0.600 & 0.657 & 0.658 & 0.651 & 0.596 & 0.744 & 0.661 & 0.742 \\
GRNet\protect~\cite{xie2020grnet} & 0.860 & 0.565 & 0.647 & 0.643 & 0.671 & 0.568 & 0.668 & 0.718 & 0.667\\ 
PMP-Net\protect~\cite{wen2021pmp} & \textbf{0.901} & 0.521 & 0.566 & 0.636 & 0.734 & 0.551 & 0.630 & 0.728 & 0.658\\ \midrule
Ours & 0.842 & 0.622 & 0.660 & \textbf{0.671} & \textbf{0.764} & \textbf{0.603} & \textbf{0.750} & \textbf{0.759} & \textbf{0.709}\\ \bottomrule
\end{tabular}

\caption{Quantitative comparison results with other methods of point cloud completion on ShapeNet using F-Score (threshold is 0.01, higher is better). Note that the metric is computed by 16,384 points.}
\label{tab:shapenetf1}
\end{table}

\subsection{Implementation Details}
We train the feature extractor in the pretext task by implementing a PointNet++ including three set abstraction layers. Then we apply two fully connected layers after feature extractor to classify each complete point cloud into the corresponding label with the cross-entropy loss. The input of the pretext task is a set of complete point clouds $P$ with  coordinates and object category labels. 
To avoid the large domain gap between different datasets, we chose 8 categories of complete point clouds from the train set of ShapeNet as the dataset for pretext task. We train the feature extractor with batch size 64 until the convergence and utilize the feature space before fully connected layers as the representation of global geometry shapes.

For completion network (the bottom branch of Fig.\ref{fig:fig1}), we only input partial point clouds from ShapeNet, Completion3D, and KITTI dataset with its coordinate information. 
Besides, we apply the farthest point sampling (FPS) algorithm at the end of partial-to-sparse process to subsample points and preserve the global structure of sparse output. 
We utilize the Adam optimizer to train the whole architecture of point cloud completion in 150 epochs with batch size 16 and learning rate $1e-4$ on two GTX 2080Ti GPUs.

\begin{table*}[htb!]
\setlength{\tabcolsep}{2.5mm}
\centering
\begin{tabular}{l|cccccccc|c}

\toprule 
Methods/CD-$\ell_{2}(\times10^{-4})$$\downarrow$ & Airplane & Cabinet & Car & Chair & Lamp & Sofa & Table & Vessel & Overall \\ 
\midrule
FoldingNet\protect~\cite{yang2018foldingnet} & 12.83 & 23.01 & 14.88 & 25.69 & 21.79 & 21.31 & 20.71 & 11.51 & 19.07\\
PCN\protect~\cite{yuan2018pcn} & 9.79 & 22.70 & 12.43 & 25.14 & 22.72 & 20.26 & 20.27 & 11.73 &  18.22\\
PointSetVoting\protect~\cite{zhang2020point} & 6.88 & 21.18 & 15.78 & 22.54 & 18.78 & 28.39 & 19.96 & 11.16 & 18.18 \\
AtlasNet\protect~\cite{groueix2018papier} &  10.36 & 23.40 & 13.40 & 24.16 & 20.24 & 20.82 & 17.52 & 11.62 & 17.77 \\
TopNet\protect~\cite{tchapmi2019topnet} & 7.32 & 18.77 & 12.88 & 19.82 & 14.60 & 16.29 & 14.89 & 8.82 & 14.25  \\
SoftPoolNet\protect~\cite{wang2020softpoolnet} & 4.89 & 18.86 & 10.17 & 15.22 & 12.34 & 14.87 & 11.84 & 6.48 & 11.90 \\
SA-Net\protect~\cite{wen2020point} & 5.27 & {\color{red}14.45} & {\color{red}7.78} & 13.67 & 13.53 & 14.22 & 11.75 & 8.84 & 11.22\\
GRNet\protect~\cite{xie2020grnet} & 6.13 & 16.90 & {\color{blue}8.27} & 12.23 & 10.22 & 14.93 & {\color{blue}10.08} & 5.86 & 10.64\\
PMP-Net\protect~\cite{wen2021pmp} &{\color{red}3.99} & {\color{blue}14.70} & 8.55 & {\color{blue}10.21} & {\color{blue}9.27} & {\color{red}12.43} & {\color{red}8.51} & {\color{blue}5.77} & {\color{red}9.23}\\
\midrule
\textbf{Ours} &{\color{blue}4.1} & 15.72 & 8.39 & {\color{red}9.84} & {\color{red}8.43} & {\color{blue}13.75} & 11.82 & {\color{red}5.58} & {\color{blue}9.76}\\ 
\bottomrule
\end{tabular}
\caption{Quantitative comparison with state-of-the-art methods on Completion3D using Chamfer Distance (CD, lower is better) with L2 norm multiplied by $10^{4}$. Note that the metric is computed by 2,048 points. The best results are marked in {\color{red}Red} and the second best results are marked in {\color{blue}Blue}.}
\label{tab:all-com3d}
\label{tab:completion}
\end{table*}

\section{Comparison with State-of-the-art Methods}
\subsection{Quantitative Comparison.}
Table~\ref{tab:shapenetcd} and ~\ref{tab:shapenetf1} show the quantitative comparison results of previous methods and ours on ShapeNet.
It is noteworthy that output of MSN~\cite{liu2020morphing} contains 8,192 points while others have 16,384 points per object. We concatenate two times inferred results for fair comparison.

Experimental results show that our method outperforms others on all metrics, which indicates that our method has a powerful ability to reconstruct complete point clouds.
Besides, we evaluate our method on Completion3D benchmark. Specifically, we utilize the network pre-trained on ShapeNet and test on Completion3D. We subsample the output 16,384 points from pre-trained network to 2,048 for evaluation. Table~\ref{tab:completion} shows the comparison results with other methods in which our method achieves the best performance in terms of average Chamfer Distance. 

Since we don't have the ground truth of KITTI dataset, we utilize the consistency metric proposed in ~\cite{yuan2018pcn}. Specifically, the consistency calculates the average CD between each frame of the same tracklet. This measurement describe the robustness of the network for different partial inputs of the same object.
\begin{equation}
    \begin{aligned}
        Consistency =\frac{1}{T}\sum_{t}^{T}\left( \frac{1}{F_t-1}(\sum_{f=1}^{F_t-1}CD(\hat{P}_{f}, \hat{P}_{f+1}))\right),
    \end{aligned}
\end{equation}
where $T$ denotes the number of tracks or objects, $F_{t}$ denotes the number of frames for track $t$.
Table~\ref{tab:kitti} shows the comparison results between our method and GRNet. Experimental results show that our method is more powerful against the variations in the partial input.

\begin{table}[]
\centering
\begin{tabular}{l|c}
\toprule 
Methods & Consistency($\times 10^{-3}$)$\downarrow$~~ \\ 
\midrule
AtlasNet\protect~\cite{groueix2018papier}  & 0.700\\
PCN\protect~\cite{yuan2018pcn} & 1.557\\
FoldingNet\protect~\cite{yang2018foldingnet} & 1.053\\
TopNet\protect~\cite{tchapmi2019topnet} & 0.568\\
MSN\protect~\cite{liu2020morphing} & 1.951\\
GRNet\protect~\cite{xie2020grnet} & 0.313\\
PMP-Net\protect~\cite{wen2021pmp} & 0.472\\\midrule
Ours& \textbf{0.295}\\
\bottomrule
\end{tabular}
\caption{Quantitative comparison with other state-of-the-art methods on KITTI using Consistency metric (lower is better) mutiplied by $10^{3}$. }
\label{tab:kitti}
\end{table}

\begin{table}[t]
\setlength{\tabcolsep}{0.6mm}
\centering
\begin{tabular}{c|cccc|c|c}
\toprule
Methods & SPF & Pri & DS & Proj & CD-$\ell_{1}(\times10^{-3})\downarrow$ & F-Score@1$\%\uparrow$ \\ \midrule
A & - & - & - & - & 9.16 & 0.635 \\
B & 1 & \checkmark & - & - & 8.58 & 0.652 \\
C & 2 & \checkmark & - & - & 8.42 & 0.660 \\
D & 3 & \checkmark & - & - & 8.39 & 0.662 \\
E & 3 & \checkmark & Cos & - & 8.10 & 0.692\\
F & 3 & - & Cos & \checkmark & 8.24 & 0.682\\
G & 3 & \checkmark & L2 & \checkmark & 8.12 & 0.695\\
H & 3 & \checkmark & Cos & \checkmark & \textbf{8.05} & \textbf{0.709}\\
\bottomrule
\end{tabular}
\caption{Ablation study on ShapeNet dataset. We investigate different designs including SPF module (SPF) with or without shape priors (Pri), difficulty-based sampling scheme (DS) and projection loss (Proj).}
\label{tab:ablation}
\end{table}

\subsection{Qualitative Comparison.}
We compare the qualitative performance of our method with other state-of-the-art methods. 
Fig.~\ref{fig:compare}(a) shows the illustration of completion results on ShapeNet. 

Specifically, the visualization on the first row shows that all of methods succeed in completing the global shape of a standard airplane. However, as for some point clouds in non-standard geometric shapes, previous methods mostly tend to generate mean shape with blurry details or standard shape. The visualization results on the second and third row show that FoldingNet, PCN, and MSN fail to generate local geometric details of shape information. Also, results on the bottom row show that FoldingNet, PCN, GRNet and PMP-Net tend to predict standard chairs with four legs although the ground truth does not have legs. Results predict by PMP-Net have lots of noisy points.
Our method outperforms these methods on completing these non-standard details.

Besides, we also show the comparison results of point cloud completion on KITTI dataset which is illustrated in Fig.~\ref{fig:compare}(b). We implement the completion using the network which is fine-tuned under car category on ShapeNet dataset for both ours and GRNet.
Experimental results show that our method is more powerful to predict the geometric details of cars, especially for some extreme sparse point clouds, which are shown on the top two rows. 



\section{Ablation Study}
\label{sec:ablation}
We conduct comprehensive ablation studies to evaluate the importance of different components in our method: soft-perceptual priors, SPF module, difficulty-based sampling strategy and projection loss. 
Without loss of generality, we conduct all experiments on ShapeNet dataset. The results are summerized in Table~\ref{tab:ablation}. The baseline A is the vanilla GRNet trained without Gridding loss\cite{xie2020grnet}.

\noindent\textbf{SPF and Soft-perceptual Priors.} 
To demonstrate the effectiveness of the SPF module and soft-perceptual priors, we evaluate our method with different numbers of SPF modules on multi-scale (Method B, C, D, gradually added from low-level to high-level). Besides, we also demonstrate how the soft-perceptual priors affect the performance of the framework (Method F, H). Specifically, SPF without soft-perceptual priors means that we discard the prior $u$ in Fig.~\ref{fig:spf}. Without the prior, our proposed SPF module can fuse local and global features from low to high, and helps to preserve detail geometric information. And experimental results show that priors learned from our proposed prototype-aware heterogeneous task can maximize the effect.

\noindent\textbf{Difficulty-based Sampling.} 
We also evaluate the effectiveness of difficulty-based sampling strategy with different metrics. Table~\ref{tab:ablation} shows the evaluation results of whether using re-weight strategy with different metrics or not. We can see that re-weight strategy helps to improve the performance of point cloud completion (Method D, E). Besides, comparison results (Method G, H) show that using cosine similarity has better performance than L2 norm. That is because the former metric is more robust to point clouds in different scale or densities while the latter is more sensitive.

\noindent\textbf{Projection Constraint.}
Table~\ref{tab:ablation} shows the quantitative results of whether using projection constraint $\mathcal{L}_{proj}$ for point clouds (Method E, H). Experimental results show that 2D constraint makes improvement for point cloud completion. That is because it can effectively encourage the network to preserve the boundary details. Furthermore, the constraint on sparse and dense point clouds helps to drop out some noisy points in the partial-to-sparse process.

\begin{figure}[]
    \centering
    \includegraphics[width=\linewidth]{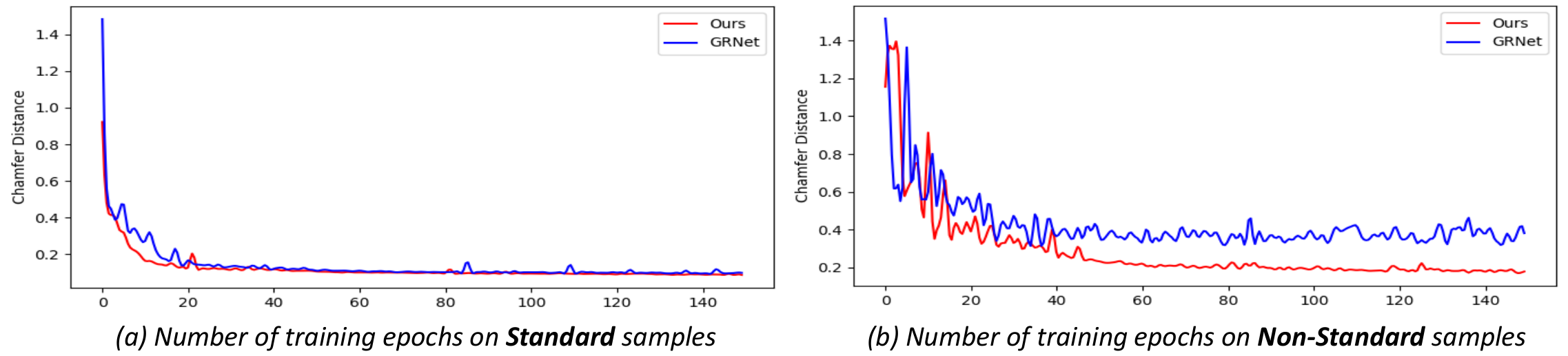}
    \caption{Average Chamfer Distance metric as function of training epochs on standard (Left) and non-standard (Right) shapes for both GRNet (Baseline) and our method.}
    \label{fig:abcomplete}
\end{figure}

\noindent\textbf{Completion of Standard and Non-standard Samples.}
Fig.~\ref{fig:abcomplete} shows the average Chamfer Distance metric of the same batch of standard and non-standard objects of airplane category in validation set as function of epochs during training process using GRNet and our method.
Specifically, we rank the feature distance of partial point clouds to the prototypical center from near to far and take the top 20$\%$ and the last 20$\%$ as standard and non-standard samples, respectively.
Compared with GRNet ({\color{blue}Blue}), our method ({\color{red}Red}) performs well on standard samples while improving the performance on non-standard samples.

\section{Conclusion}
In this work, we propose a prototype-aware heterogeneous task to learn soft-perceptual priors for the following point cloud completion task. To make the prior enhance the completion  performance on non-standard samples, we propose the SPF module which fuses the prior with both global and local features. Besides, to tackle the variety of missing parts in different partial point clouds, we propose a  difficulty-based sampling scheme guided by the perceptual feature gap between each pair of partial and complete objects. Experimental results show that our method has a more powerful ability in predicting geometric shapes with both global and local details, especially for some non-standard objects. 


\begin{figure*}[]
    \centering
    \includegraphics[width=\linewidth]{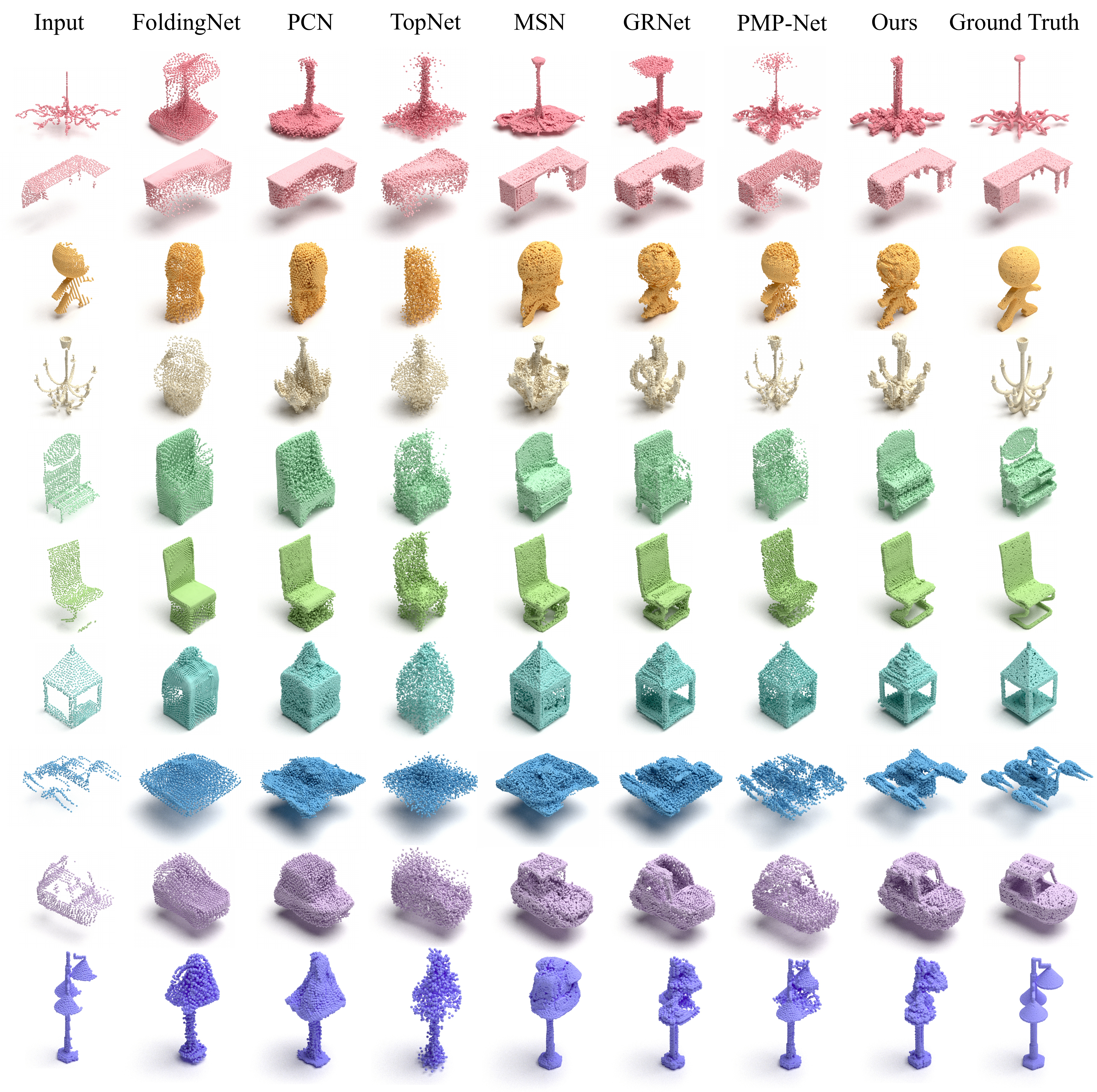}
    \caption{Illustration of point cloud completion comparison with previous methods on different categories of ShapeNet dataset. For ShapeNet, from left to right columns are the partial input, completed results from FoldingNet\protect\cite{yang2018foldingnet}, PCN\protect\cite{yuan2018pcn}, TopNet\protect\cite{tchapmi2019topnet}, MSN\protect\cite{liu2020morphing}, GRNet\protect\cite{xie2020grnet}, PMP-Net\protect\cite{wen2021pmp}, our method and the last is the ground truth.}
    \label{fig:morecompare}
\end{figure*}

{\appendix[More Quantitative Results]

We present more qualitative comparison results of point cloud completion on ShapeNet dataset. Fig.~\ref{fig:morecompare} shows the extra visualization results of our method and other four methods on ShapeNet. Experimental results show that our method has the best performance on completing the local geometric details.

}


\bibliographystyle{IEEEtran}
\bibliography{reference}

\vspace{11pt}




\vfill

\end{document}